\documentclass[letterpaper, 10 pt, conference]{ieeeconf}
\IEEEoverridecommandlockouts

\usepackage{amsfonts}
\usepackage{amsmath}
\usepackage{graphicx}
\usepackage{amssymb}
\usepackage{mathtools}
\usepackage{xcolor}
\usepackage{caption}
\usepackage{subcaption}
\usepackage{float}
\usepackage{stmaryrd}
\usepackage{epstopdf}
\usepackage{multirow}
\usepackage{color,soul}
\usepackage{accents}
\usepackage{float}
\usepackage{verbatim}
\usepackage{algorithm} 
\usepackage{mathrsfs}  
\usepackage{caption}
\usepackage{subcaption}
\usepackage{tabularx}
\usepackage{xcolor}
\usepackage{placeins}
\usepackage{comment}
\usepackage{tabularray}

\usepackage{mathtools}
\usepackage{algpseudocode} 

\allowdisplaybreaks

\usepackage{tabularray}
\newcolumntype{b}{X}
\newcolumntype{C}[1]{>{\centering\let\newline\\\arraybackslash\hspace{0pt}}m{#1}}

\newtheorem{theorem}{Theorem}[section]

\newtheorem{assumption}{Assumption}
\newtheorem{corollary}[theorem]{Corollary}

\newtheorem{definition}[theorem]{Definition}

\newtheorem{remark}[theorem]{Remark}
\newtheorem{problem}[theorem]{Problem}

\title{\LARGE \bf
Spatiotemporal Tube-Based Safety-Certificate for Autonomous Navigation of Articulated Vehicles}
\author{Mohd. Faizuddin Faruqui$^{1,3}$, Ratnangshu Das$^{1}$, Ravi Kumar L$^{3}$, and Pushpak Jagtap$^{1,2}$
\thanks{$^1$Department of Cyber-Physical Systems and $^2$Department of Aerospace Engineering, Indian Institute of Science, Bengaluru, India, and $^3$U.R. Rao Satellite Centre, ISRO, Bangalore, India. {\tt\small \{ratnangshud,pushpak\}@iisc.ac.in}, {\tt\small \{mfaruqui,rkkumarl\}@ursc.gov.in}.}%
}

\begin{document}
\maketitle
\thispagestyle{empty}
\pagestyle{empty}

\begin{abstract}
Articulated vehicles are the workhorses of freight transportation, and their autonomous navigation is challenging. Their physical characteristics and motion constraints pose significant challenges in maneuvering these vehicles on narrow routes. This paper presents a spatiotemporal tube-based approach to plan autonomous navigation of vehicles like tractor-semi-trailers, truck/ tractor trailers, towing Automated Guided Vehicles (AGVs), and road trains. This planning approach provides a certified path plan for the truck or tractor, ensuring that the towed series of trailers always remains within the road corridor, limited by permissible corrections. The planning leverages the kinematics of the linked elements along with sway constraints to arrive at a safe tube for the actuated prime mover. We modify the spatiotemporal tube using permissible corrections to provide a route safety certificate to the vehicle for the given route. The proposed planning method is verified on a truck-trailer navigation simulation for a complex route.

\end{abstract}


%

\section{Introduction}
Large trucks accounted for 9.6\% of vehicles involved in fatal crashes in the United States, while heavy goods vehicles were involved in 14.2\% of fatal crashes in Europe, primarily due to human error.
Autonomous navigation of these vehicles with safety guarantees can mitigate these safety risks. Various aspects of safe operations of these vehicles are studied in \cite{he2023review}, \cite{van2024safe}, \cite{regehr2009safety}. 
In addition, existing highways are not used optimally due to precautionary safety margins to accommodate human errors. Automated Highway Systems (AHS) offer intelligent infrastructure for autonomous guidance and navigation of vehicles. Articulated vehicles also have extensive applications in  intralogistics \cite{fottner2021autonomous} in warehouse and factory settings and construction and mining/ quarrying \cite{dragt2005overview} setups. A combination of high load capacity and modularity with superior maneuverability and terrain adaptability makes articulated vehicles indispensable across industries with demanding operations.   

An optimization-based safety-critical trajectory planning method for truck-trailer vehicles is discussed in \cite{gao2025efficient,7139245}.  
State estimation method of particle filter for motion planning of articulated vehicles with active trailer steering is discussed in \cite{10253020}. Motion planning for autonomous robots primarily uses methods based on stochastic planning, like Rapidly-exploring Random Tree (RRT) \cite{ljungqvist2019motion}, \cite{7759544}, and Probabilistic RoadMaps (PRM) \cite{4338880}, which by itself cannot provide safety guarantees. Path planning for an articulated vehicle with $N$ trailers on a constrained road/corridor is a complex problem due to kinematic coupling and non-holonomic restrictions. Reactive path planning approaches for safe motion of robots are reviewed in \cite{rahimi2020review}. Kinodynamic motion planning that employs geometric sampling followed by funnel-based control to track the planned trajectory is discussed in \cite{9911990}. These methods rely on an accurate vehicle model and on runtime feasibility that can only guarantee conditional safety, rather than an upfront certificate that guarantees safety a priori for a given route.  

\par  Recent progress in Advanced Driver Assistance Systems (ADAS) has provided several safety features for these vehicles, but most of these safety mechanisms are reactive and may not provide formal guarantees. To achieve guaranteed safety, this paper proposes planning for autonomous control of articulated vehicles for freighting on complex routes. The underactuated kinematics of articulated vehicles pose several challenges in motion planning, including jackknifing and deadlocks at sharp and narrow turns. We propose a Spatiotemporal Tubes (STT)-based approach \cite{10540043,das2025spatiotemporal} to plan safe motion of articulated vehicles, such as a truck/ tractor with multiple trailers, and provide a safety certificate for the route suitability of an articulated vehicle in a particular configuration. We consider the problem of maneuvering a prime mover with multiple trailers autonomously using STT-based planning, which certifies safety by guaranteeing that the vehicle remains safe as long as it is confined within its respective synthesized STTs. We do this by using the STT of the prime mover and the vehicle kinematic map to find unsafe instances of the last towed trailer tube and then arrive at a corrected tube for the prime mover so that the whole vehicle remains safe. This planned tube can also serve as a temporary certificate for highway usage, enabling safe and optimal use of the road capacity. Safety during long-haul freight transportation using heavy articulated vehicles is an essential requirement of the goods supply chain industry. 
\par  The key highlight of our method is planning under unknown detailed dynamics of the vehicle and then arriving at a route safety certificate for a particular configuration of the vehicle.  
\section{Notation and Preliminaries}
The symbols $\mathbb{R}, \mathbb{R}_0^+, \mathbb{R}^+, \mathbb{Z}^+, \mathbb{Z}_0^+, \mathbb{S}^1$, and $\mathbb{S}^2_{++}$ denote sets of reals, non-negative reals, positive reals, positive integers, non-negative integers, unit circle, and $2 \times 2$ symmetric positive definite matrices, respectively. The symbol $u^\perp$ represents an orthogonal component of the vector $u$ expressed as $u^\perp = R(90^0)u$, $u\in \mathbb{R}^2$ and $R(90^0)$ is the rotation matrix corresponding to a $90^0$ counter-clockwise rotation. The Euclidean norm of a vector $x$ is denoted by $\|x \|$  and the Frobenius norm of a matrix $X$ is denoted by $\|X \|_F$. A column vector with $n$ rows is represented as $\mathbb{R}^n$, and the vector space with $m$ rows and $n$ columns is represented as $\mathbb{R}^{m \times n}$, where $n,m \in \mathbb{Z}^+$. The symbol $\oplus$ denotes the Minkowski sum such that for two sets $A,B \subset \mathbb{R}^n$, the Minkowski sum is $A \oplus B = \lbrace a+b:a\in A,b\in B \rbrace$. The notation $C^1(\cdot,\cdot)$ denotes the space of continuously differentiable functions, where the first argument denotes the domain and the second argument the co-domain. The symbol $\bigcirc_{i=0}^{N} \kappa_i(\cdot)$ represents the successive composition in order given by $\bigcirc_{i=0}^{N} \kappa_i(\cdot) := \kappa_N(\kappa_{N-1}(\cdots\ \kappa_0(\cdot))$. The expression $u.v$ denotes the dot product between two vectors $u\in \mathbb{R}^n$ and $v\in\mathbb{R}^n$ expressed as $u^\top v$, where $u^\top$ denotes the transpose of the vector $u$. Given an index set $\mathcal{I} = \lbrace 0,1,2,...,N \rbrace$ of $N+1$ subsystems, where $N \in \mathbb{Z}^+$ and sets $X_i$, $i \in \mathcal{I}$, the Cartesian product of these sets is given by $X =\prod_{i\in \mathcal{I}} X_i:= X_0 \times X_1 \times \cdots \times X_N = \lbrace (x_0,x_1,\ldots,x_N)|x_i \in X_i, i\in \mathcal{I}\rbrace$.
\subsection{System Description}
\label{SecIIa}
The articulated vehicle considered in this paper is a truck/tractor, hereafter referred to as the prime mover, with $N$ serially connected trailers. This can be classified as a kinematically-coupled interconnected system with $N+1$ subsystems composed as $\Sigma= \lbrace \Sigma_0, \cdots,\Sigma_N \rbrace$. The kinematics of the $i^{th}$ trailer vehicle is represented as $\Sigma_i$, $i \in \mathcal{I}\setminus\lbrace0\rbrace$, and the kinematics of the prime mover vehicle is denoted by $\Sigma_0$. The interconnection among subsystems is captured by the composition maps $\Omega$ and $\Xi$ defined in Section \ref{PbmFor}. Since all the towed subsystems/trailers are passive, they are indirectly influenced by the prime mover input through inter-subsystem kinematic coupling. 

The interconnected system consists of subsystems joined by a hitch-hinge mechanism, with the prime mover subsystem leading the chain. To each subsystem $i$, the follower subsystem $i+1$ is connected by a rod of length $L$, and the relative sway angle of the hinge between the subsystems is $\delta_i(t)$ for all $i \in \mathcal{I} \setminus \{N\}$, where $\delta_i(t)$ is constrained as $\delta_i(t) \in [-\theta_i,\theta_i]$ and $\theta_i$ is the maximum sway angle determined by physical limitation of the hinge.

 \par The state space $X_i$ of each subsystem $\Sigma_i, i\in \mathcal{I}$ in the serially interconnected system $\Sigma$ is closed and connected. The prime mover and trailers are constrained to planar articulation with $p_i(t) \in \mathbb{R}^2$ as the position of the $i^{th}$ subsystem, such that $p_i(t) = \pi(x_i(t))$ and $\pi:X_i \rightarrow \mathbb{R}^2$ is the position projection map. The overall state space of the interconnected system $\Sigma$ is expressed by the Cartesian product $X \subset \mathbb{R}^{2(N+1)} \times (\mathbb{S}^1)^{N+1}$.

\section{Problem Formulation}

\label{PbmFor}
Autonomous navigation of kinematically-coupled interconnected systems is a complex task and reactive approaches cannot guarantee safety. In this study, we consider synthesizing a confinement zone in the form of an STT for the prime mover and subsequently for $N$ trailers using kinematic coupling, such that the complete articulated vehicle remains
within the given route at all times, as shown in Figure \ref{Ellip_Geom}. The problem considered in this paper includes the prime mover towing $N$ trailers, each assumed to have one axle. The axle articulation angles are limited, and hence, inter-subsystem collision scenarios $p_i(t) = p_j(t), i \neq j$, do not arise. Obstacle avoidance task is out of scope of this paper but can be incorporated in the STT itself using tube-circumvent as discussed in \cite{10540043}, \cite{11187220}. In this paper, we use STTs with ellipsoidal cross-sections, which we call Spatiotemporal Ellipsoidal Tubes (STETs). Ellipsoidal tubes are efficient tools for safety-critical control design in robust control and reachability analysis due to their quadratic form expression and Linear Matrix Inequalities (LMI) computations \cite{kurzhanski2000ellipsoidal,jeantube}. STET is defined as follows:
\begin{definition}[Spatiotemporal Ellipsoidal Tube (STET)]
\label{PTSTET}
A Spatiotemporal Ellipsoidal  (STET) for a subsystem $i\in \mathcal{I}$ is a time-varying, set-valued non-degenerate ellipsoidal bound $\mathcal{E}_i(q_i(t),Q_i(t))$ for all $t\in [0,T]$, where $T\in \mathbb{R}^+$ is finite time horizon over which STET is defined. It is parameterized by $q_i(\cdot) \in C^1([0,T], \mathbb{R}^2)$ and $Q_i(\cdot) \in C^1([0,T], \mathbb{S}_{++}^{2})$ which are functions representing the center of the ellipsoid and the shape matrix of the ellipsoid respectively for all time $t \in [0,T]$. The diagonal elements of $Q_i$ quantify axis-aligned scaling, and the off-diagonal elements capture the orientation of the ellipsoid. STET for a subsystem $i$ for all $i\in \mathcal{I}$ is given by
\begin{equation}
    \label{EllIneq}
    \hspace{-1em}\mathcal{E}_i(q_i(t)\!,\!Q_i(t))\! = \!\lbrace \!\psi_i(t)\! \in\! \mathbb{R}^2 | \psi^\top_i(t) Q_i(t) \psi_i(t)\!\leq\! 1\rbrace\!\subseteq\!\mathbb{R}^2,\!
\end{equation}    
where $\psi_i(t)=p_i(t)-q_i(t)$ and $p_i(t) \in \mathbb{R}^2$ is the position of the $i^{th}$ subsystem.
\end{definition}

The outward support radius of the ellipsoid $\mathcal{E}_i(q_i(t),Q_i(t))$ along the unit vector $\hat{n} \in \mathbb{S}^{1}$ is:
\begin{equation}
\label{suprad}
    r(Q_i(t),\hat{n}) = {1}/{\sqrt{\hat{n}^\top Q_i(t) \hat{n}}}.
\end{equation}
We assume that the reference route to be tracked by the vehicle is given a priori and defined as follows:

\begin{definition}[Route Corridor]
    \label{route}
    Given the center $c(\cdot) \in C^1([0,T],\mathbb{R}^2)$ and the width profile $w(\cdot) \in C^1([0,T],\mathbb{R}^+)$, the Route Corridor $\mathcal{R}(t) \subset \mathbb{R}^2$ is a continuously differentiable time varying corridor $(\mathcal{P}_l(t),\mathcal{P}_r(t))$, for all $t\in [0,T]$, defined as
    \begin{equation}
          \hspace{-0.5em} \mathcal{R}(t) := \lbrace p \in \mathbb{R}^2 \ | \ |(p-c(t))^\top\hat{n}(t)| < 0.5w(t)\rbrace,
    \end{equation}    
    where $\hat{n}(t) \in \mathbb{S}^1$ is the unit normal direction associated with the route centerline $c(t)$ and the left and right boundaries 
    $\mathcal{P}_l(t), \mathcal{P}_r(t) \in C^1([0,T],\mathbb{R}^2)$ are respectively given by
    \begin{subequations}
        \begin{align}
            \mathcal{P}_l(t) =& c(t) + 0.5w(t).\hat{n}(t),\\
            \mathcal{P}_r(t) =& c(t) - 0.5w(t).\hat{n}(t). 
        \end{align}
    \end{subequations}
\end{definition}
Initially a nominal prime mover STET $\mathcal{E}_0(q_0(t)\!,\!Q_0(t))$ is generated using route centerline $c(t)$ and prime mover dimensions with margins. We intend the vehicle to perform an Autonomous Safe Navigation (ASNav) Task defined as follows: 
\begin{definition}[Autonomous Safe Navigation Task]
Given the kinematically interconnected system $\Sigma$ with subsystem states $x_i \in X_i$, let $p_i(t) = \pi(x_i(t))$ be the planar position of the $i^{th}$ subsystem such that $p_i(0) \in \mathcal{R}(0)$ for all $i\in \mathcal{I}$. An Autonomous Safe Navigation (ASNav) Task enforces the position trajectories of all subsystems $\Sigma_i$ such that $p_i(t) \in \mathcal{R}(t)$ for all $i\in \mathcal{I}$, for all $t \in [0,T]$.  
    \label{PTANav}
\end{definition}

Given the route $\mathcal{R}(t)$, we aim to synthesize a STET for the vehicle such that it can perform ASNav task and eventually assign a \textit{Route Safety Certificate} (RSC) $\mathscr{C}(\Sigma,\mathcal{R}(t))$ such that vehicle configuration $\Sigma$ is \textit{Safe} iff $\mathscr{C}(\Sigma,\mathcal{R}(t)) \geq 0$ else \textit{Unsafe}. 

\begin{definition} [Kinematic Map]
\label{KineMap}
    
     Consider a kinematically coupled articulated vehicle consisting of a prime mover towing $N$ trailers, represented as an interconnected system $\Sigma$. Let $\mathcal{E}_i(q_i(t),Q_i(t))$ be a STET for the $i^{th}$ subsystem such that $q_i(t) \in \mathbb{R}^2$ and $Q_i(t) \in \mathbb{S}_{++}^2$ represent center and shape matrix respectively for all $i \in \mathcal{I}, t \in [0,T]$, as described in Definition \ref{PTSTET}. For each hinge joint belonging to $i^{th}$ subsystem characterized by inter-axle length $L$ and sway angle $\delta_i(t)\in [-\theta_i,\theta_i]$, let $\kappa_i$ and $\iota_i$ represent smooth center and shape map expressed as follows 
    \begin{subequations}\
    \label{XiMaths}
        \begin{align}
        \kappa_i: &\mathbb{R}^2 \times \mathbb{S}^1 \rightarrow \mathbb{R}^{2}, q_{i+1}(t) = \kappa_i(q_i(t), \delta_i(t)), \\
        \iota_i: & \mathbb{S}^{2}_{++} \times \mathbb{S}^1  \rightarrow \mathbb{S}^{2}_{++}, Q_{i+1}(t) = \iota_i(Q_{i}(t),\delta_i(t)).   
        \end{align}
    \end{subequations}
    The center of the STET of the $(i+1)^{th}$ follower subsystem is related to its leading subsystem through the hinge kinematics as
    \begin{equation}
    \label{hingekine}
    q_{i+1}(t) = q_i(t) - L\hat{u}_i(t,\delta_i(t)),      
    \end{equation} 
    where $\hat{u}_i(t,\delta_i(t))$ is the unit vector along the interconnection link at $t \in [0,T]$, defined as
    \begin{equation}
    \label{uvec}
      \hat{u}_i(t,\delta_i(t)) = \cos\delta_i(t) \hat{e}_i(t) + \sin\delta_i(t)\hat{e}_i(t)^\perp, \forall i \in \mathcal{I}.  
    \end{equation}
    Here, $\hat{e}_i(t)$ is the velocity unit vector and $\hat{e}_i(t)^\perp = R(90)\hat{e}_i(t)$ is the corresponding planar orthogonal unit vector. The deterministic propagation of the STET centers and shape matrices from the prime mover to the last trailer along the interconnected kinematic chain is obtained by composing the corresponding center and shape maps over all the hinge joints as:
    \begin{subequations}
    \label{compo}
     \begin{align}
        &\Omega(t,\delta_0(t),\cdots,\delta_{N-1}(t)) := \bigcirc_{i=0}^{N-1} \kappa_i(\cdot,\delta_i(t)),\label{Omega}\\ 
        &\Xi(t,\delta_0(t),\cdots,\delta_{N-1}(t)):= \bigcirc_{i=0}^{N-1}\iota_i(\cdot,\delta_i(t)),\label{Xi}\\ 
        &q_N(t) = \Omega(t,\delta_0(t),\cdots,\delta_{N-1}(t))(q_0(t)),\label{Omegapp}\\
        &Q_N(t) = \Xi(t,\delta_0(t),\cdots,\delta_{N-1}(t))(Q_0(t))\label{xiapp}.
     \end{align}
    \end{subequations}
\end{definition}

For an articulated vehicle consisting of a prime mover and serially interconnected subsystems, the STETs of the follower subsystems are spatially and temporally offset from the prime mover's STET owing to inter-axle kinematic coupling and propagation delays.

    \begin{assumption}
    \label{lipsAssu}
        The composition maps $\Omega(t,\delta_0(t),\cdots,\delta_{N-1}(t))$ and $\Xi(t,\delta_0(t),\cdots,\delta_{N-1}(t))$ as defined in Equation~\eqref{Omega}-\eqref{xiapp} with the sway angle tuple defined as $\hat{\delta}(t):=(\delta_0(t),\cdots,\delta_{N-1}(t))$ 
        are Lipschitz in their respective arguments uniformly for all  $t \in [0,T]$ and $\hat{\delta}(t) \in \Theta := \prod_{i=0}^{N-1} [-\theta_i,\theta_i]$, also there exists constants $\mu^c,\mu^s \in \mathbb{R}^+$ such that
    \begin{subequations}
    \label{Lischitz}
        \begin{align}
            &\|\Omega(t,\hat{\delta}(t))(q) - \Omega(t,\hat{\delta}(t))(q')\| \leq  \mu^c \|q - q'\|,\\
            &\|\Xi(t,\hat{\delta}(t))(Q) - \Xi(t,\hat{\delta}(t))(Q')\|_F \leq  \mu^s \|Q-Q'\|_F, 
        \end{align}
    \end{subequations}
    for all $q,q' \in \mathbb{R}^2$ and  $Q,Q' \in \mathbb{S}^2_{++}$.
    \end{assumption}

The problem statement considered in this paper is expressed formally as follows:
\begin{problem}
\label{problem}
    Given the composition of interconnected subsystems $\Sigma= \lbrace \Sigma_0, \dots,\Sigma_N \rbrace$ with serial kinematic relations parameterized by inter-axle length $L$ and sway angles $\delta_i$ as defined in Definition~\ref{KineMap} and a prescribed route corridor $\mathcal{R}(t)$ defined over finite time horizon $t \in [0,T]$ as in Definition~\ref{route}, consider that Assumption~\ref{lipsAssu} hold. The objective is to provide a \emph{Route Safety Certificate} $\mathscr{C}(\Sigma,\mathcal{R}(t))$ for the vehicle system $\Sigma$ and the given route $\mathcal{R}(t)$. 
    \label{Pbmstmt}
\end{problem}

\begin{remark}
We solve the problem by constructing a safe STET $\mathcal{E}_0(t)$ of the prime mover for all $t \in [0,T]$. In case the safe certificate $\mathscr{C}(\Sigma,\mathcal{R}(t))$ exists, the prime mover STET $\mathcal{E}_0(t)$ should be such that the articulated vehicle satisfies the ASNav task as stated in Definition \ref{PTANav}.
\end{remark}

\section{PATH PLANNING }
\begin{figure}[ht!]
	\centering	\includegraphics[width=1\linewidth,scale= 0.5]{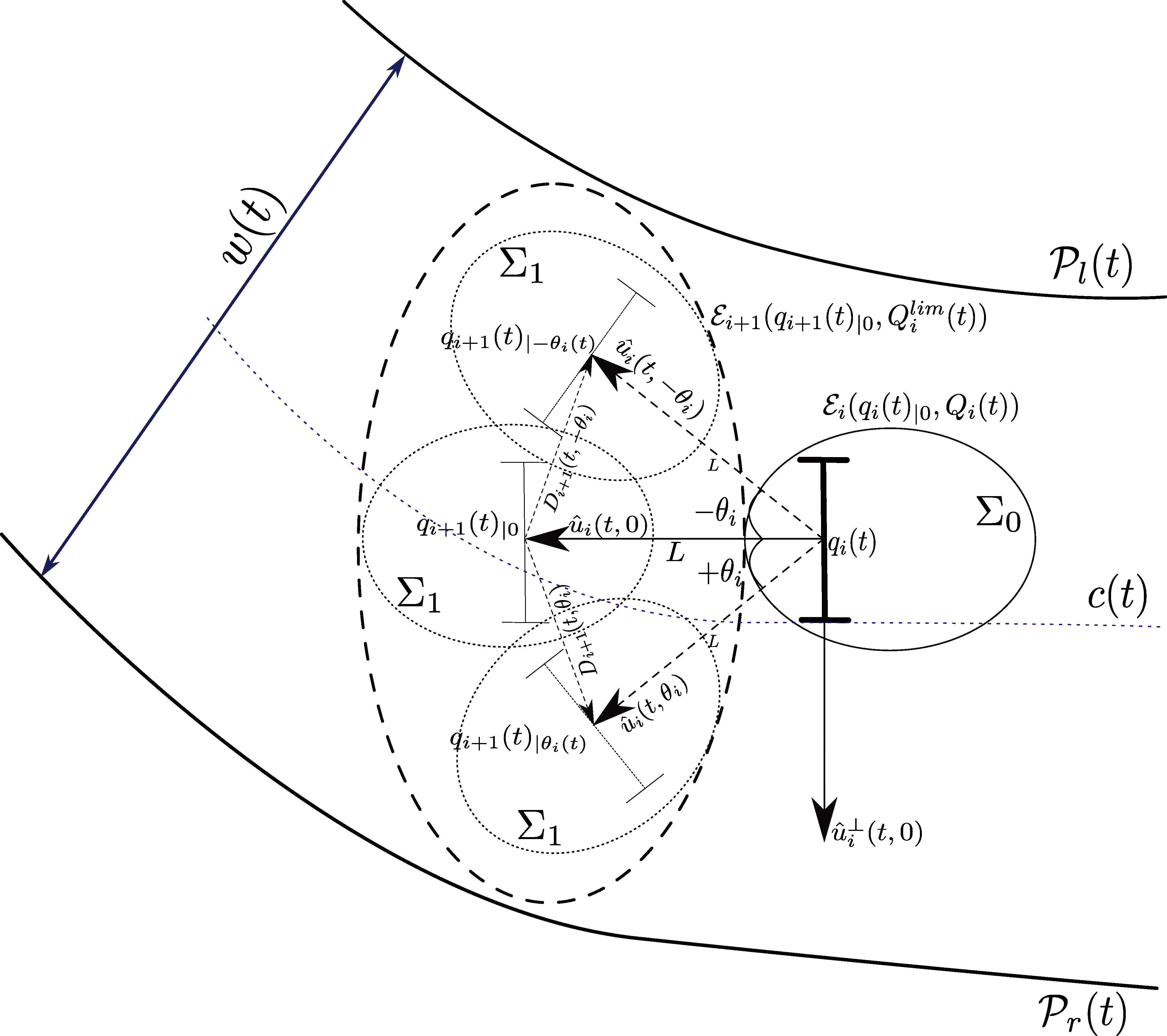}
	\caption{Kinematic geometry for an instant of prime mover position}
	\label{Ellip_Geom}
\end{figure}

The path planning problem here now becomes finding the STET of the prime mover such that all trailers remain confined to the route $\mathcal{R}(t)$. We use the kinematic geometry as illustrated in Figure \ref{Ellip_Geom} to find the feasible tubes for the towed trailers. 
Each towed trailer can sway by an angle $\delta_i(t) \in [-\theta_i, \theta_i]$ relative to its leading vehicle hinge. Hence, for each hinge point state $p_i(t) \in \mathcal{E}_i$ of the leading vehicle, the follower axle center state $p_{i+1}(t)$ satisfies $p_{i+1}(t) \in \mathcal{E}_{i+1}(t)$. Let $\mathcal{M}_i(t)$ denote the reachable set of the follower vehicle induced by the tube $\mathcal{E}_i(t)$ and sway angle $\delta_i(t)$. The set $\mathcal{M}_i(t)$ is expressed as
\begin{equation}
\label{M_i}
\mathcal{M}_i(t)\! =\! \mathcal{E}_i(t) \oplus \mathcal{H}_i^{\delta_i}(t), \delta_i(t)\!\in \! [-\theta_i,\theta_i], \forall i \in \mathcal{I}, \forall t\!\in\![0,T],    
\end{equation}
where $\mathcal{E}_i(t)$ is the ellipsoidal region where the leader vehicle can take positions and $\mathcal{H}^\delta_i(t)$ accounts for all follower's displacements due to $\delta_i(t)$ sway articulation relative to the leader's position in $\mathcal{E}_i(t)$. $\mathcal{H}^\delta_i(t)$ is the set of all displacement vectors from $q_i(t)$ to $q_{i+1}(t)$ relative to the leader's hinge sway $\delta_i(t)$ and kinematic constraint at time $t$, it is given as 
\begin{equation}
    \mathcal{H}_i^{\delta_i}(t) = \lbrace -L\hat{u}_i(t,\delta_i(t)) | \delta_i(t)\in [-\theta_i,\theta_i] \rbrace,
\label{Ri}
\end{equation}
where $\hat{u}_i(t,\delta_i)$ is the unit vector along the inter-axle link direction. Consider the prime mover STET $\mathcal{E}_0(q_0(t),Q_0(t))$ as defined in \eqref{EllIneq}. The displacement of the $(i+1)^{th}$ vehicle's center due to $\delta_i(t)$ sway at the hinge of the $i^{th}$ vehicle is expressed as
\begin{equation}
    D_{i+1}(t,\delta_{i}(t)) = q_{i+1}(t)_{|\delta_{i}(t)} - q_{i+1}(t)_{|0},
    \label{displa}
\end{equation}
using the kinematic relation illustrated in Figure \ref{Ellip_Geom} and Equation~\eqref{hingekine} and \eqref{uvec}, this can be written as 
\begin{equation}
       D_{i+1}(t,\delta_{i}(t))= L((1-\cos\delta_{i}(t))\hat{u}_{i}(t,0) - \sin\delta_{i}(t)\hat{u}^\perp_{i}(t,0)).
       \label{disp}
\end{equation}
The peak sways in the longitudinal and lateral directions corresponding to extreme sway angles $\delta_i = \pm \theta_i$ are
  \begin{align}
    a_i:=  L(1-\cos\theta_i),\    b_i:=  L\sin\theta_i.
    \label{peak}
    \end{align}
Using the peak deviations along unit vectors $\hat{u}_i \in \mathbb{R}^2$ and $\hat{u}_i^\perp \in \mathbb{R}^2$, we define the shape matrix $Q_i^{lim}(t)$ as follows
\begin{equation}
    Q_i^{lim}(t) = U_i \begin{bmatrix}
        1/a_i^2 & 0 \\ 0 & 1/b_i ^2 
    \end{bmatrix} U_i^\top, \quad U_i = [\hat{u}_i(t,0),\hat{u}_i^\perp(t,0)]^\top.
    \label{Qlim}
\end{equation}

The tube of the follower vehicle, centered at $q_{i+1}(t)_{|0}$, is constructed using Equation~\eqref{M_i} and the Young/ Kurzhanski-Varaiya bound  \cite{10.5555/646880.710463, 8619508, kurzhanski2000ellipsoidal}, considering centers at zero as
\begin{equation}
    \mathcal{E}_{i}(0,Q_i(t)) \oplus \mathcal{E}_i^{lim}(0,Q_i^{lim}(t))  \subseteq \mathcal{E}_{i+1}(0,Q_i^b(\beta,t)),
    \label{delta0}
\end{equation}
\begin{equation}
    Q_i^b(\beta,t) = [(1+\frac{1}{\beta})Q_i^{-1}(t) + (1+\beta)(Q_i^{lim}(t))^{-1}]^{-1},
    \label{Qib}
\end{equation}
where $\beta > 0$ and its optimal value $\beta^{o}$ \cite{8619508} is given as
\begin{equation}
    \beta^{o} = \sqrt{\frac{trace((Q^{lim}_{i}(t))^{-1})}{trace(Q_i^{-1}(t))}}, \forall t\in[0,T].
    \label{betaeqn}
\end{equation}
Using Equations \eqref{delta0} and \eqref{Qib}, the STET of the $(i+1)^{th}$ follower subsystem is given by $\mathcal{E}_{i+1}(q_{i+1}(t)_{|0},Q_i^b)$. This ellipsoid serves as a conservative outer bound on the Minkowski sum of the sway-induced ellipsoids associated with $\delta_i(t)$. In particular, it encloses the sum of the sway-limiting ellipsoid $\mathcal{E}_i^{lim}(0,Q_i^{lim}(t))$ hinged at the $i^{th}$ leader, and the ellipsoid $\mathcal{E}_i{(0,Q_i(t))}$ within which the leader hinge $q_i(t)$ may lie.
\section{Safe path plan certificate for a route}
In this section, we arrive at a safety certificate for an articulated vehicle in a particular configuration on a given route. For a multi-trailer configuration consisting of a prime mover with each body represented by a STET, we deduce an algorithm, as stated in Algorithm \ref{safecert}, that adjusts the center and shape matrix of the prime mover's ellipsoid based on violations by the last trailer, and plans the center and shape matrix of the prime mover such that all follower trailers are safe. To enable the algorithm, we define \emph{Boundary leeway} as a measure of clearance of the tube ellipsoid of the last trailer from the route boundary as follows: 
\begin{definition}[Boundary leeway] 
    Given the route $\mathcal{R}(t)\subset \mathbb{R}^2$, confined by $C^1$ curves representing the left boundary $\mathcal{P}_l(t)$ and the right boundary $\mathcal{P}_r(t)$ with respect to the direction of motion, let $\hat{n}_l(x)$ and $\hat{n}_r(x)$ denote the unit inward normals for all $x(t)\in \mathcal{P}_l(t)$ and $\mathcal{P}_r(t)$, respectively. Given a subsystem ellipsoid $\mathcal{E}(q(t),Q(t))$ with center $q(t)$ at time $t$, we define the left and right boundary leeway as:
    \begin{equation}
    \label{leeway}
        \begin{aligned}
        \eta_l = \zeta_l - r_l, \eta_r = \zeta_r - r_r,\\
        \end{aligned}
    \end{equation}
    respectively, where $r_l(Q(t),\hat{n}_l(t))$ and $r_r(Q(t),\hat{n}_r(t))$ are the support radii, as defined in Equation~\eqref{suprad}, in the left and right directions, respectively. The boundary offsets $\zeta_l$ and $\zeta_r$ are defined as
    \begin{equation}
    \label{Centre_dist}
        \begin{aligned}
        \zeta_l \!=\! \inf_{x \in \mathcal{P}_l} (x-q(t)).\hat{n}_l(x), \zeta_r \!=\! \inf_{x \in \mathcal{P}_r} (x - q(t)).\hat{n}_r(x).\! 
        \end{aligned}
    \end{equation}
\end{definition}
To meet the ASNav task specifications, we use the center-shift and ellipsoid-shrink approach to ensure that all trailers remain confined to the route. We track violations of the last trailer ellipsoid and accordingly plan center-shift and ellipsoid-shrink for the prime mover. The permissible corrections applied to the prime mover are then defined as follows.  
\begin{definition}[Permissible corrections]
\label{permcorr}
Given the Lipschitz continuous composition maps $\Omega:q_0(t) \mapsto q_N(t)$ and $\Xi: Q_0(t) \mapsto Q_N(t)$  for all $t\in[0,T]$, the following  corrections to the prime mover tube ellipsoid are permissible:
\begin{itemize}
    \item Permissible center corrections $\Delta q_0(t)$ satisfying $\| \Delta q_0(t) \| \leq \Delta q^{max}_{0}$, such that $\tilde{q}_0(t) = q_0(t) + \Delta q_0(t)$ implemented in Algorithm $CORRCEN$ (\ref{centcorr1}), and
    \item Permissible ellipsoid shrink $\tilde{Q}_0(t) = Q_0(t)/\gamma_0(t)^2$ with $\gamma_0(t) \in [\gamma_0^{min},1], \gamma_0^{min} \in (0,1]$ implemented in Algorithm $SHRINK$ (\ref{shapecorr}).
\end{itemize}
Under these permissible corrections, the composition maps $\Omega$ and $\Xi$ remain  Lipschitz on their respective domains, as in Assumption~\ref{lipsAssu}, with modified constants $\tilde{\mu}^c, \tilde{\mu}^s
\in \mathbb{R}^+$ for the center and shape maps, respectively.

\label{ShrShp}    
\end{definition}
Next, we establish the existence of safe corrected tubes.
\begin{theorem}
Given the route $\mathcal{R}(t)\subset \mathbb{R}^2$, confined by $C^1$ curves representing the left boundary $\mathcal{P}_l(t)$ and the right boundary $\mathcal{P}_r(t)$ with reference to the direction of motion, consider an articulated vehicle modeled as an interconnected system $\Sigma$ with $N$ trailers and kinematic maps as defined in Definition~\ref{KineMap}. Let $\rho(t) = \min (\eta_l(t), \eta_r(t))$ denote the worst-side boundary leeway. To satisfy the ASNav task specifications, let $\Delta q_0(\cdot)$ and $\gamma_0(\cdot)$ represent permissible center correction and ellipsoid shrinkage of the prime mover, respectively, as defined in Definition~\ref{ShrShp}. If, for all $t \in [0,T]$, the worst-side boundary leeway satisfies
\begin{equation}
    \rho(t) \geq \tilde{\mu}^c \Delta q_0^{max} + \sup_{\hat{n}\in \mathbb{S}^1}(r(Q_N(t),\hat{n}) - r(\tilde{Q}_N(t),\hat{n})),
    \label{bound}
\end{equation}
then there exist valid correction terms $\Delta q_0(\cdot)$ and $\gamma_0(\cdot) \in [\gamma_0^{min},1]$ such that the corrected STET of the last trailer remains fully contained within the route corridor for all time:
\begin{equation}
    \mathcal{E}_N(\tilde{q}_N(t),\tilde{Q}_N(t)) \subset \mathcal{R}(t), \forall t.
    \label{LastCondn}
\end{equation}
Here, $r(\cdot,\hat{n})$ denotes the support radius along the direction $\hat{n}$ and $\tilde{\mu}^c$ is the modified Lipschitz constant introduced in Definition~\ref{ShrShp}.
As a result, the articulated vehicle satisfies the ASNav task, and a \emph{Route Safety Certificate} (RSC) $\mathscr{C}(\Sigma, \mathcal{R}(t))$ exists.
\label{RSCExist}
\end{theorem}
Due to space limitations, the proof of the theorem will be included in an expanded version as a journal submission.
  \begin{algorithm}[H]
	\caption {\text{ROUTE\_CERT}$(\Sigma,\mathcal{R}(t),\Omega,\Xi$)} 
    \label{safecert}
	\begin{algorithmic}[1]
    \Require Route corridor $\mathcal{R}(t):\lbrace \mathcal{P}_l(t),\mathcal{P}_r(t), w(t) \rbrace $, Kinematic maps $\Omega$ and $\Xi$, prime mover STET $\mathcal{E}_0(q_0(t),Q_0(t))$ for all time $t\in [0,T]$ 
    \Ensure Route Safety Certificate (RSC) $\mathscr{C}$, safe prime mover center $\tilde{q}_0(t)$ and shape matrix $\tilde{Q}_0(t)$
    \State Initialize $Safe \leftarrow false$, $Unsafe \leftarrow false$
    \For  {time index {$i = 1,2, \ldots $}}
        \State $q_0 \leftarrow q_0(i), q_N \leftarrow q_N(i)$  \Comment{Center of the prime mover and the last trailer respectively at $i^{th}$ instant}
        \State $Q_0 \leftarrow Q_0(i), Q_N \leftarrow Q_N(i)$ \Comment{Shape matrix of the prime mover and the last trailer at $i^{th}$ instant}
        \State Compute the unit vector $\hat{e}_n$ in the direction of motion
        \State Compute left normal $\hat{n}_l = \hat{e}^\perp_n$ and the right normal $\hat{n}_r = -\hat{n}_l$
        \State Compute projections $\zeta_l$ and $\zeta_r$ on $\mathcal{P}_l$ and $\mathcal{P}_r$, resp. 
        \State Compute support radius $r_l$ and $r_r$ along normals $\hat{n}_l$ and $\hat{n}_r$ respectively
        \State Compute boundary leeway $\eta_l$ and $\eta_r$ along normals $\hat{n}_l$ and $\hat{n}_r$ respectively      
        \If {$\min(\eta_l,\eta_r) \leq 0$}
            \State $\tilde{q}_0 \leftarrow CORRCEN(\eta_l,\eta_r,\Omega)$ \Comment{Shift prime mover center opposite to last trailer violation side by $\Delta q_0$ as in Definition \ref{permcorr}} \label{centcorr1}
            \State Propagate $\tilde{q}_0$ and $Q_0$ through the kinematic map $\Omega$ and $\Xi$ to get $\tilde{q}_N$ and $\tilde{Q}_N$ respectively, then calculate $\eta_l$ and $\eta_r$ again
            \If {$\min(\eta_l,\eta_r) \leq 0$}
                \State $\tilde{Q}_0\leftarrow
                SHRINK(Q_0,\eta_l,\eta_r)$ \Comment{If violation of last trailer persists post center correction, isotropic-shrink prime mover tube as in Definition \ref{permcorr}} \label{shapecorr} 
                    \State                 
                    $\tilde{q}_0(i) \gets \tilde{q}_0, \tilde{Q}_0(i) \gets \tilde{Q}_0$
            \Else
                \State $\tilde{q}_0(i) \gets \tilde{q}_0, \tilde{Q}_0(i) \gets Q_0$  
            \EndIf
        \Else
            \State $\tilde{q}_0(i) \gets {q}_0, \tilde{Q}_0(i)\gets Q_0$
        \EndIf
    \EndFor
        \State $\mathscr{C}(\Sigma,\mathcal{R}(t)) \leftarrow CERT(\tilde{q}_0(\cdot),\tilde{Q}_0(\cdot),\Omega,\Xi,\mathcal{R}(\cdot))$ \label{certif}  
        \If {$\mathscr{C}(\Sigma,\mathcal{R}(t)) \geq 0$}
            \State $Safe \leftarrow true$
            \Else
                \State $Unsafe \leftarrow true$
        \EndIf
\end{algorithmic}
\end{algorithm}
\textbf{Summary:} If the route is wide enough relative to: \\
(i) the vehicle geometry: Larger trailers will require a larger support radius $r(Q_N(t),\hat{n}_\sim)$, \\
(ii) the trailer chain length: Longer chains imply a larger Lipschitz constant of center propagation $\tilde{\mu}^c$, \\
(iii) the allowed sway: Larger sway limits $\delta_i(t)\in [-\theta_i,\theta_i]$ means larger reachable sets and thus larger ellipsoids that again increase $r(Q_N(t),\hat{n}_\sim)$, and \\
(iv) the permitted corrections: Maximum center shift $\Delta q^{max}_{0}$ and Ellipsoid shrink factor $\gamma_0(t)$, then we can always adjust the prime mover slightly, and the entire articulated vehicle remains safe. $\rho(t)$ basically quantifies the runtime safety with respect to route boundaries as
\begin{equation}
\!\rho(t)
\!\ge\!
\underbrace{\! \tilde{\mu}^c \Delta q_0^{\max}}_{\substack{\text{Trailer chain length} \\ \text{and center correction}}}
\!+\!
\underbrace{\!
\sup_{\hat{n}\in\mathbb{S}^1}
\bigl(
r(Q_N(t),\hat{n}) - r(\tilde{Q}_N(t),\hat{n})
\bigr)
}_{\substack{\text{Vehicle geometry,} \\ \text{sway limits and shrinkage}}}.
\end{equation}

\begin{corollary}
Theorem \ref{RSCExist} establishes that the permissible center and shape corrections exist and are bounded above by the worst-side boundary leeway $\rho(t)$. For a given route $\mathcal{R}(t)$ and the kinematic composition maps $\Omega$ and $\Xi$, we derive a scalar Route Safety Certificate (RSC) $\mathscr{C}(\Sigma,\mathcal{R}(t))$ from the lower bound of $\rho(t)$. 

Consider the route $\mathcal{R}(t) \subseteq \mathbb{R}^2$, and let $c(t)$ and $w(t)$ denote the given route center and width, respectively, for all $t\in [0,T]$. Given the composition maps $\Omega$ and $\Xi$, let $h_{\Omega}, h_{\Xi} \in \mathbb{R}_0^+$ be the constants such that the following bounds hold:\\
(i) the normal displacement between the corrected prime mover center and the corrected last trailer center is bounded by $h_{\Omega}$:
$| (\tilde{q}_N(t) - \tilde{q}_0(t)).\hat{n}_c (t)| \leq h_{\Omega}, \forall t \in [0,T]$,
(ii) the maximum support radius of the last trailer ellipsoid is bounded by $h_{\Xi}$ times the maximum support radius of the prime mover ellipsoid:
$r_N^{max} \leq h_{\Xi}r_0^{max}.$

    Here $\hat{n}_c(t)$ is the normal to the direction of motion, and $r_i^{max}(t) = \sup_{\|\hat{n}\|=1} 1/\sqrt{\hat{n}^\top\tilde{Q}_i(t)\hat{n}}$ denotes the support radius of the $i^{th}$ ellipsoid arrived using Equation~\eqref{suprad} for all $t \in [0,T]$ and let $r_i^{max}:=\sup_{t \in [0,T]}r_i^{max}(t)$. 
    
    Using these bounds, we define a lower bound on the worst-side boundary leeway, termed the \textit{Route Safety Certificate}, as the difference between the minimum route half width $w_{min}:= \min_{\mathrm{t\in[0,T]}} (0.5w(t))$, and the sum of the kinematic offset $h_\Omega$ and the maximum trailer radius $r_N^{max}$. Specifically, the RSC is implemented in Algorithm $CERT$ (\ref{certif}) and is given by
    \begin{equation}
        \mathscr{C}(\Sigma,\mathcal{R}(t)) : = w_{min} - (h_\Omega + r_N^{max}).
    \end{equation}
    The RSC provides a binary safety assessment of the vehicle–route pair:
\begin{subequations}
\begin{align}
    \mathscr{C}(\Sigma,\mathcal{R}(t)) \geq 0 & \implies Safe, \\
    \mathscr{C}(\Sigma,\mathcal{R}(t)) < 0  & \implies Unsafe.
\end{align}
\end{subequations}
A nonnegative RSC $\mathscr{C}(\Sigma,\mathcal{R}(t)) \geq 0$ implies that the corrected prime-mover tube, when propagated through the kinematic chain, ensures that the last trailer remains fully contained within the route corridor, i.e., Equation~\eqref{LastCondn} is satisfied. Conversely, a negative RSC $\mathscr{C}(\Sigma,\mathcal{R}(t)) <0$ indicates that the permissible corrections defined in \ref{permcorr} are insufficient to guarantee safety, and therefore the route is unsafe for the articulated vehicle configuration $\Sigma$.
\end{corollary}
\begin{figure*}[!t]
	\centering
	\includegraphics[width=0.99\linewidth]{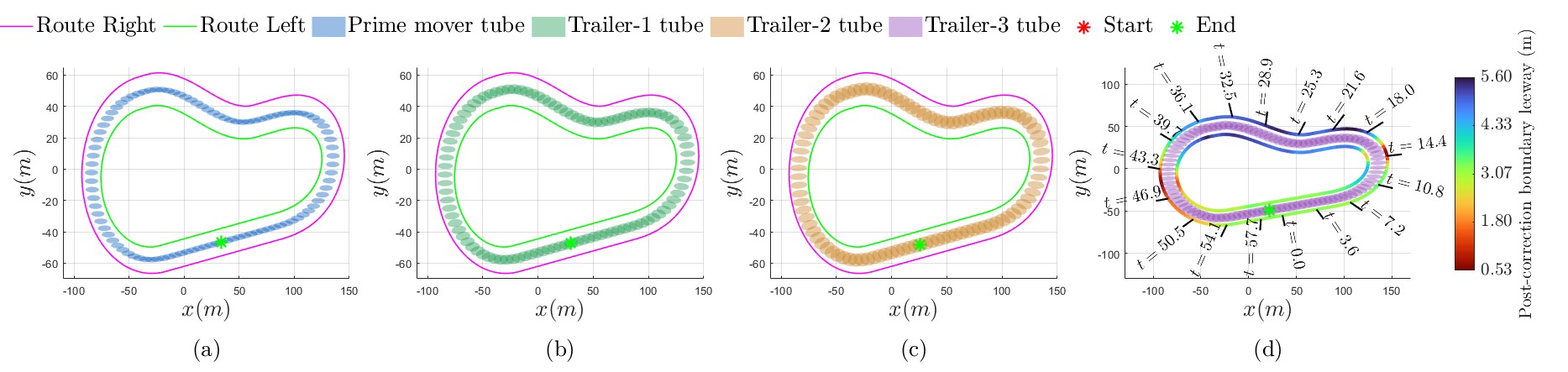}
	\caption{STETs for ASNav task in scenario-5 (a) Prime mover tube, (b)-(d) Trailer-1 to Trailer-3 tubes (color-map along left and route boundary of trailer-3 represents boundary leeway)}
	\label{Alltubes}
\end{figure*}

\begin{figure*}[!t]
    \centering
    \begin{subfigure}[t]{0.31\textwidth}
        \centering
        \includegraphics[width=0.945\linewidth]{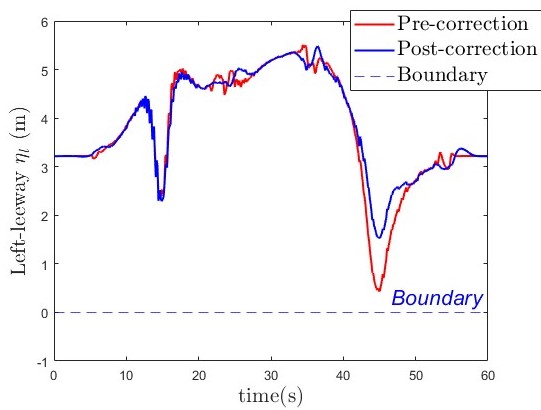}
        \caption{}
        \label{Left_Lee}
    \end{subfigure}
    \hfill
    \begin{subfigure}[t]{0.31\textwidth}
        \centering
        \includegraphics[width=0.945\linewidth]{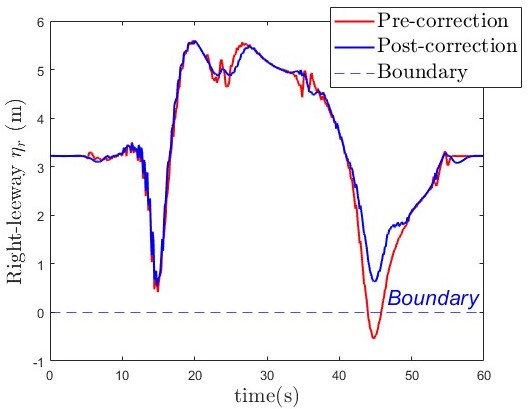}
        \caption{}
        \label{Right_Lee}
    \end{subfigure}
    \hfill
    \begin{subfigure}[t]{0.31\textwidth}
        \centering
        \includegraphics[width=0.945\linewidth]{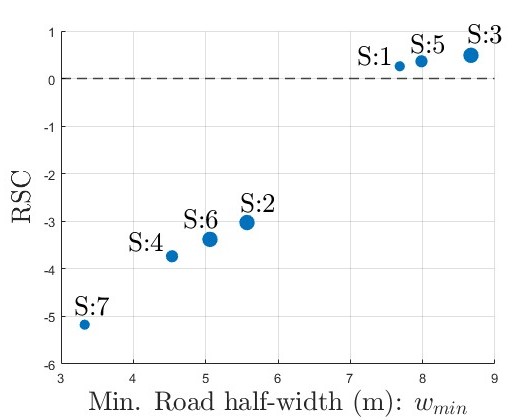}
        \caption{}
        \label{RSC_SCENE}
    \end{subfigure}
    \caption{Simulation results (route length 543.620 m): (a) Left leeway $\eta_l$ for last trailer in scenario-5, (b) Right leeway $\eta_r$ for last trailer scenario-5, (c) RSC for scenarios 1-7 (S:1 to S:7) against $w_{min}$}
    \end{figure*}

\section{Simulations}
We prove the effectiveness of our planning approach using a truck as a prime mover towing multiple trailers. To test our planning method, we consider multiple scenarios, as listed in Table~\ref{Comp2_table}, configured using different parameters, $\textit{viz.}$, the number of subsystems $N+1$, hitch length $L$, sway limit $\pm \theta$, trailer dimension $\mathcal{D} (length \times width)$, bounds $h_\Omega$ and $h_\Xi$, and the minimum route half width $w_{min}$. The STETs for the prime mover and trailers for Scenario 5 are shown in Figure~\ref{Alltubes}, and the boundary leeway before and after permissible corrections on the left and right sides are shown in Figure~\ref{Left_Lee} and \ref{Right_Lee}, respectively. Figure \ref{RSC_SCENE} shows the RSC values for the different scenarios plotted against the narrowest instant of route half width $w_{min}$. It can be observed that scenarios $1,3$ and $5$ are safe with respect to the given route corridor and vehicle configuration. All scenario simulations were run on a computer with an Intel i7-12650H and 32 GB RAM. RSC generation time scaled linearly with the number of trailers, with $2$ trailers taking around $0.5s$ and $5$ trailers taking around $1.15s$.

\begin{table}[t]
\centering
\caption{Route and vehicle configuration scenarios}
\label{Comp2_table}
\scriptsize
\resizebox{\columnwidth}{!}{%
\begin{tblr}{
  colspec = {c c c c c c c c},
  row{1} = {font=\bfseries},
  cell{1-Z}{1-Z} = {c,m},
  hlines,
  vlines
}
Sc. No. & $N+1$ & $L$ (m) & $\pm\theta$ (deg.) & $\mathcal{D}$ (m $\times$ m) & $h_\Omega$ & $h_\Xi$ & $w_{\min}$ (m) \\
\textbf{1} & 3 & 4.67 & 9.4  & 11.2 $\times$ 2.6 & 0.231 & 1.003 & 7.683 \\
\textbf{2} & 6 & 3.54 & 11.0 & 11.3 $\times$ 2.8 & 0.636 & 1.004 & 5.572 \\
\textbf{3} & 6 & 3.97 & 10.7 & 10.1 $\times$ 2.7 & 0.335 & 1.020 & 8.668 \\
\textbf{4} & 4 & 3.50 & 5.9  & 12.2 $\times$ 2.8 & 0.351 & 1.001 & 4.535 \\
\textbf{5} & 4 & 4.19 & 12.4 & 10.1 $\times$ 2.5 & 0.250 & 1.006 & 7.984 \\
\textbf{6} & 6 & 4.91 & 5.4  & 10.9 $\times$ 2.7 & 0.534 & 1.005 & 5.059 \\
\textbf{7} & 3 & 4.13 & 9.6  & 11.6 $\times$ 2.7 & 0.426 & 1.002 & 3.328
\end{tblr}
}
\end{table}

\begin{table*}[!t]
\label{Comp_table}
    \centering
    \caption{Comparison of safety assurance mechanisms for articulated vehicles within a given route corridor}
    \label{Comparison_Table}
    \begin{tblr}{
      colspec={|X[0.32]|X[0.82]|X|X|}, row{1} = {c}, 
      hlines,
      vlines,
      cell{1}{1-4} = {c,m},   
      cell{2-Z}{1-4} = {m},       
      row{1} = {font=\bfseries}}
      \textbf{Feature} & \textbf{Proposed- RSC corridor certification} & \textbf{Optimization-based approach and efficient planner \cite{OLIVEIRA202015572}, \cite{11159493}} & \textbf{Robust planning \cite{7989693}} \\
        Guarantee tool   &  Route Safety Certificate (RSC) that decides for a vehicle corridor safety   &   Feasibility based safety: Sequential Quadratic Programming applied to a non-linear optimal control problem &  Robust Control Invariant (RCI) tubes and Control Contraction Metrics (CCM)       \\
        System knowledge   &  Relies on kinematic bounds, hence suited for low slip regimes &  Detailed knowledge of vehicle kinematics and route details required &  Detailed non-linear dynamics of the system and disturbance characterization required \\
        Computational Efficiency  &  No heavy optimization-based calculations required   &   Computationally heavy due to trajectory optimization/ search problem  &  Performs convex optimization, hence computationally expensive      \\
        Scalability   &  Calculation is done using STET, hence linear scaling with trailers   &  Trailer count increases dimensionality, more iterations to achieve feasible solutions &  Polynomial/ exponential scaling with increase in states or degree of polynomials    \\
    \end{tblr}
\end{table*}

\subsection{Comparison}
As the planning approach proposed in this paper works with ellipsoidal bounds of the vehicle subsystems, it does not need detailed nonlinear dynamics of the system and handle bounded unknown disturbances. Standard methods that deploy optimization-based methods and other planning tools need detailed dynamics of the system. STET-driven route certificate depends only on the pose states of the vehicle, hence it is light on computational efficiency and scalability compared to other methods. Table~\ref{Comparison_Table} summarizes the features where the proposed planning framework has an edge in terms of safety assurance mechanism for articulated vehicles compared to other methods. It can be observed that the proposed method provides a fast certificate layer with moderate model requirements in contrast to \cite{OLIVEIRA202015572}-\cite{7989693}, which has strong feasibility and model dependencies.
\section{Conclusion}
In this paper, we proposed a planning problem for an articulated vehicle tackling an autonomous navigation task. The articulated vehicle considered in this paper comprised of a prime mover serially connected to trailers with kinematic coupling. The offline planning adopted in this paper using STET certifies the route corridor of the complete vehicle kinematic chain under bounded sway angles and guarantees formal safety using the RSC. 
    \par In future work, we plan to extend this planning method for multi-vehicle system on an automated highway. Multiple vehicles can be allotted spatiotemporal driven certified tubes for their proposed routes. In addition, we can use certified offline planning as a worst-case fall back feature and have online reactive controllers \cite{das2025real} to take care of unknown disturbances or events on highways for wider applicability of autonomous navigation beyond automated highways to unstructured environments.




\bibliographystyle{IEEEtran}

\bibliography{./MyLibrar/MyLibrary}

\end{document}